\documentclass{article}

\usepackage[preprint]{neurips_2026}

\usepackage{microtype}
\usepackage{graphicx}
\usepackage{booktabs}
\usepackage{amsmath,amssymb,amsthm,mathtools}
\usepackage{algorithm}
\usepackage{algorithmic}
\usepackage{enumitem}
\usepackage{hyperref}
\usepackage{url}
\hypersetup{hidelinks}
\usepackage[table]{xcolor}
\usepackage{array}
\usepackage{multirow}
\usepackage{longtable}
\usepackage{etoc}

\usepackage{booktabs}
\usepackage{multirow}
\usepackage{graphicx}
\usepackage{caption}
\usepackage{makecell}

\usepackage{tcolorbox}

\theoremstyle{definition}

\theoremstyle{plain}
\newtheorem{theorem}{Theorem}

\theoremstyle{remark}

\newcommand\samethanks{\footnotemark[\value{footnote}]}

\title{\textsc{MemoRepair}: Barrier-First Cascade Repair in Agentic Memory}
\author{
    \mdseries Yang Zhao\thanks{Equal contribution.}\and
    Chengxiao Dai\samethanks\and
    Mengying Kou\and
    Yue Xiu
  }

\begin{document}
\maketitle

\begin{abstract}
Agentic memory evolves across tasks into durable derived artifacts:
summaries, cached outputs, embeddings, learned skills, and executable
tool procedures. When a source artifact is deleted, corrected, or
invalidated by tool or API migration, descendants derived from that
source can remain visible and steer future actions with stale support.
We formalize this failure mode as the cascade update problem, where
repair targets the visible derived state of the memory store.
We present \textsc{MemoRepair}, a barrier-first cascade-repair contract
for agentic memory. A repair event induces a controlled transition from
invalidated descendant state to validated successor state: affected
descendants are withdrawn before repair, successors are constructed from
retained support and staged repaired predecessors under the current
interface, and republication is restricted to validated
predecessor-closed successors. This contract induces a scalarized
repair-selection problem for a fixed repair--cost tradeoff. We show that
the induced publication problem reduces to maximum-weight predecessor
closure and can be solved exactly by a single $s$--$t$ min-cut. 
Experiments on ToolBench and MemoryArena show that, with complete influence provenance, \textsc{MemoRepair} reduces invalidated-memory exposure from $69.8$--$94.3\%$ under systems without cascade repair to $0\%$. Compared with
exhaustive \emph{Repair all}, it recovers $91.1$--$94.3\%$ of validated
successors while reducing normalized repair-operator cost from $1.00$ to
$0.57$--$0.76$.
\end{abstract}

\section{Introduction}

Persistent memory has become a core substrate for agentic AI systems
that operate across tasks, tool calls, and sessions. In these systems,
raw interaction records and retrieved documents are transformed over time
into durable derived artifacts: summaries of prior interactions, cached
tool outputs, reusable skills, and executable tool procedures that
compress past experience into state the agent can reuse
\citep{park2023generative,shinn2023reflexion,packer2023memgpt,xu2025amem}.
These artifacts support long-horizon behavior by letting the agent act
from accumulated context rather than reconstructing each decision from
scratch.

As derived state accumulates, a source update is no longer a local edit.
When a source artifact is deleted, corrected, or invalidated by tool or
API migration, descendants derived from that source may remain visible
and continue to steer future actions with stale support. A deleted
preference may survive in a summary, a corrected tool response may remain
in a cached output, and an obsolete API contract may persist in a stored
call-chain procedure. We call this the \emph{cascade update problem}: a
root update induces repair obligations over the visible descendant state
of the memory store. The repair target is therefore the visible derived
state of agentic memory, not only the invalidated root artifact.

Prior work studies how agent memories are constructed, organized,
retrieved, and reused
\citep{park2023generative,packer2023memgpt,xu2025amem}, while tool-use
systems make persistent traces, cached outputs, API
arguments, and executable procedures concrete forms of agent state
\citep{qin2023toolllm,guo2024stabletoolbench,patil2024gorilla}.
Provenance and dependency-tracking methods explain how derived state
depends on source inputs
\citep{buneman2001prov,green2007provenance,cheney2009provenance}.
What remains missing is a post-update transition for agentic memory:
after an invalidation event, the system must decide which descendants to
withdraw, which successors can be constructed from retained support under
the current interface, and which validated successors may re-enter
service.

We introduce \textsc{MemoRepair}, a barrier-first repair contract for
agentic memory. After a deletion, correction, or migration, it first removes
the affected cascade from service, then constructs successor artifacts from
post-event valid support, repaired predecessors, and the current interface.
Only validated successors whose required predecessors are also repaired may
be republished. This yields a cost-aware selection problem: recover useful
successors without exhaustively repairing every candidate. For a fixed
repair--cost tradeoff parameter $\lambda$, the valid selections form a
predecessor-closed optimization problem, which we reduce exactly to a single
$s$--$t$ min-cut.

Experiments on ToolBench and MemoryArena cover deletion, correction, and
migration events~\citep{qin2023toolllm,guo2024stabletoolbench,
memoryarena2026}. Across cascade-unaware memory systems, $92.4$--$99.7\%$ of
post-event actions still depend on invalidated information, showing that
ordinary update and retrieval do not provide cascade-level withdrawal. Under
complete influence provenance, \textsc{MemoRepair} instead withdraws affected
memory before repair and republishes only validated successors. It preserves
nearly the same validated repairs as exhaustive \emph{Repair all} while
executing substantially less repair work. Ablations show that the min-cut
selector improves over greedy selection and that store-level repair
complements parameter-level neural repair.

\paragraph{Contributions.}
We make the following contributions:
\begin{itemize}[leftmargin=*]
\item We formalize the cascade update problem in agentic memory. A root
deletion, correction, or migration induces repair obligations over the
visible descendant state of summaries, cached outputs, prompt skills,
chain procedures, and neural skills.

\item We define \textsc{MemoRepair}, a barrier-first cascade-repair
contract. The contract specifies the visibility, support, dependency,
and validation conditions under which an affected descendant may leave
the withdrawn state and re-enter service as a successor.

\item We derive the repair-selection rule induced by this contract. For
a fixed repair--cost tradeoff, valid republication forms a
predecessor-closed selection problem, yielding an exact $s$--$t$ min-cut
solver for the scalarized objective.

\end{itemize}

\section{Method}
\label{sec:method}

\begin{figure*}[!ht]
    \centering
    \includegraphics[width=\linewidth]{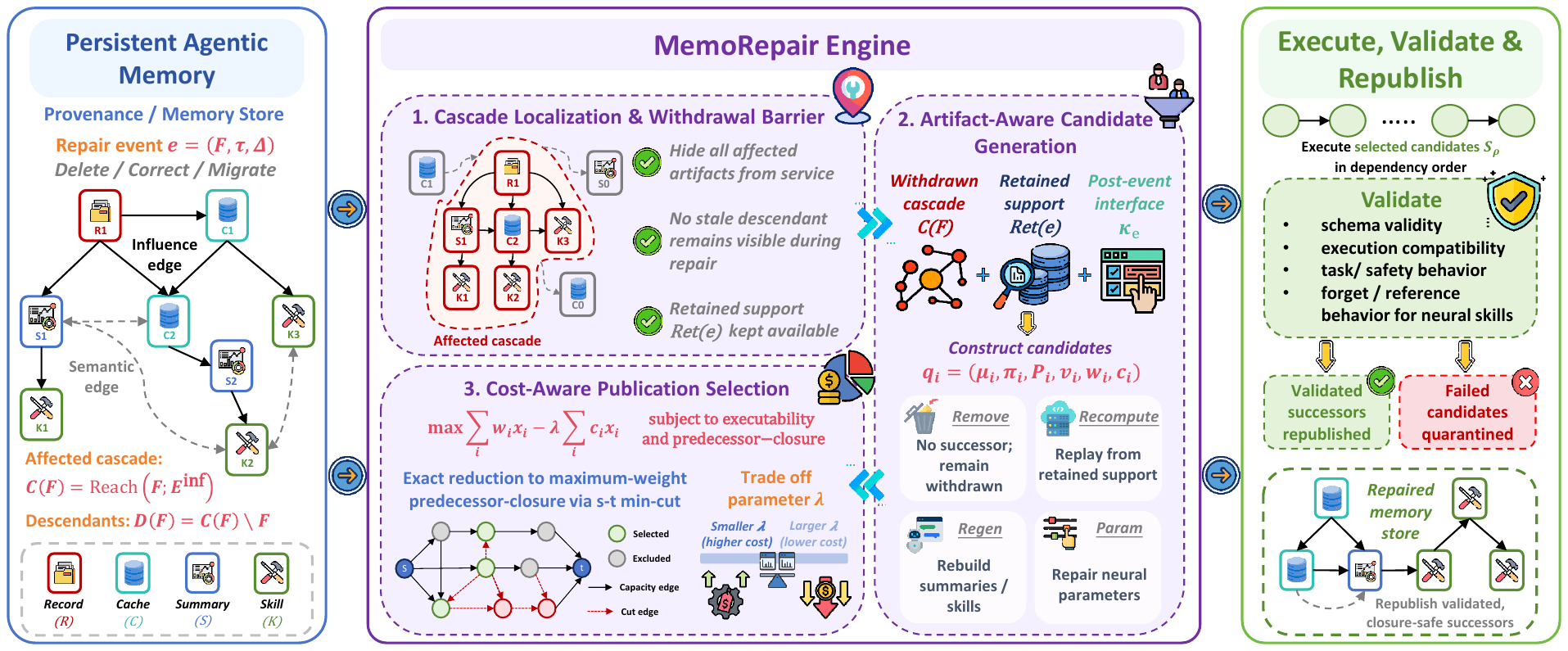}
    \caption{Repair transition enforced by \textsc{MemoRepair}: affected
    artifacts are withdrawn, successor versions are constructed, and
    validated predecessor-closed successors are republished.}
    \label{fig:framework}
\end{figure*}

A repair event in \textsc{MemoRepair} is a barrier-first transition over the
affected cascade induced by an invalidated root set. As shown in
Figure~\ref{fig:framework}, the system first withdraws the cascade from
service, constructs repair candidates for affected descendants, selects a
predecessor-closed subset under a repair--cost tradeoff, and republishes only
validated successors. 


\subsection{Problem Setup}
\label{subsec:problem-setup}

Persistent agentic memory is represented by a directed provenance graph
\[
\mathcal{G}=(V,E^{\mathrm{inf}},E^{\mathrm{sem}}),
\]
where nodes are durable artifacts. An influence edge
$(u,v)\in E^{\mathrm{inf}}$ means that $v$ was produced using $u$ as
causal support; these edges determine repair scope. Semantic edges
$E^{\mathrm{sem}}$ support retrieval only. Each artifact has a kind tag
\[
\mathrm{kind}(x)\in
\{\mathsf{record},\mathsf{cache},\mathsf{summary},\mathsf{skill}\},
\]
and each skill has an architecture tag
\[
\mathrm{arch}(x)\in
\{\mathsf{neural},\mathsf{prompt},\mathsf{chain}\}.
\]

A repair event is
\[
e=(F,\tau,\Delta),
\qquad
F\subseteq V,\quad
\tau\in\{\mathsf{delete},\mathsf{correct},\mathsf{migrate}\},
\]
where $F$ is the invalidated root set and $\Delta$ contains correction or
migration information. The affected cascade is computed only through
influence edges:
\[
C(F)=\mathrm{Reach}(F;E^{\mathrm{inf}}),
\qquad
D(F)=C(F)\setminus F ,
\]
where $\mathrm{Reach}$ includes the zero-hop roots in $F$. Thus
withdrawal targets $C(F)$, while repair candidates are generated for
descendants $D(F)$.

Let $\kappa_e$ denote the post-event interface. The retained valid support
is
\[
\mathsf{Ret}(e)=
\begin{cases}
V\setminus C(F), & \tau=\mathsf{delete},\\[2pt]
(V\setminus C(F))\cup X_\Delta, & \tau=\mathsf{correct},\\[2pt]
V\setminus C(F), & \tau=\mathsf{migrate},
\end{cases}
\]
where $X_\Delta$ is the set of fresh replacement artifacts materialized
for correction events. Thus $\mathsf{Ret}(e)$ is interpreted in the post-event artifact universe. 


\subsection{Withdrawal Barrier}
\label{subsec:withdrawal-barrier}

For the repair plan $\rho$ associated with event $e$, \textsc{MemoRepair}
initializes
\[
B_\rho\leftarrow C(F).
\]
All artifacts in $B_\rho$ are non-servable during repair. A fresh
successor $z_i$ for artifact $i$ may be republished only if
\[
\mathrm{Validate}_i(z_i)=1 .
\]
Validation checks replay consistency for recomputation, schema and
task-regression behavior for regeneration, sandbox behavior for chain
skills, and forget/reference behavior for parametric repair.

\subsection{Artifact-Aware Candidate Construction}
\label{subsec:candidate-construction}

Roots in $F$ are handled by the event semantics: deletion removes them,
correction supplies replacements, and migration updates $\kappa_e$.
For each descendant $i\in D(F)$, a deterministic mode map assigns
\[
\mu_i=\Phi_\mu^e(i;\mathcal{G},\mathsf{Ret}(e),\kappa_e)
\in
\{\mathsf{remove},\mathsf{recompute},\mathsf{regen},\mathsf{param}\}.
\]
Replayable deterministic artifacts are assigned $\mathsf{recompute}$;
summaries, prompt skills, and chain procedures are assigned
$\mathsf{regen}$ when retained support suffices; neural skills are
assigned $\mathsf{param}$ when provenance yields forget/reference
partitions. Otherwise, the artifact is assigned $\mathsf{remove}$.

For each non-removed descendant,
\[
q_i=(\mu_i,\pi_i,P_i,v_i,w_i,c_i),
\]
where $\pi_i$ is the repair operator, $P_i\subseteq D(F)$ is the set of
affected descendants whose repaired successors are consumed by $\pi_i$,
$v_i\in\{0,1\}$ records executability, and $w_i,c_i\ge0$ are fixed
value--cost terms. Root replacements and interface updates are supplied
through $\mathsf{Ret}(e)$ and $\kappa_e$, not through $P_i$.

Define the non-removed candidate index set
\[
D^+(F)=\{i\in D(F):\mu_i\neq\mathsf{remove}\},
\qquad
Q_\rho=\{q_i:i\in D^+(F)\}.
\]
For each $i\in D^+(F)$, let
\[
\bar P_i=P_i\cap D^+(F).
\]
If $P_i\setminus D^+(F)\neq\varnothing$, then $i$ requires a
non-repairable affected descendant and is marked non-executable
($v_i=0$). The repair-time dependency graph is
\[
E^{\mathrm{req}}
=
\{(j,i):i\in D^+(F),\,j\in\bar P_i\}.
\]
An edge $(j,i)$ means that the successor of $j$ must be staged before
repairing $i$. Thus $E^{\mathrm{req}}$ captures repair-time prerequisites,
not raw provenance.

\subsection{Cost-Aware Publication Selection}
\label{subsec:cost-aware-selection}

For each $i\in D^+(F)$, let $x_i\in\{0,1\}$ indicate whether $q_i$ is
selected. Define
\[
\mathrm{Repair}(x)=\sum_{i\in D^+(F)}w_i x_i,
\qquad
\mathrm{Cost}(x)=\sum_{i\in D^+(F)}c_i x_i .
\]
For a fixed tradeoff parameter $\lambda\ge0$, \textsc{MemoRepair} solves
\begin{align}
\max_x\quad
& \mathrm{Repair}(x)-\lambda\,\mathrm{Cost}(x)
\label{eq:scalarized-selection}\\
\text{s.t.}\quad
& x_i\le v_i,
\quad \forall i\in D^+(F),
\label{eq:avail}\\
& x_i\le x_j,
\quad \forall i\in D^+(F),\ \forall j\in\bar P_i,
\label{eq:closure}\\
& x_i\in\{0,1\},
\quad \forall i\in D^+(F).
\label{eq:binary}
\end{align}
Constraint~\eqref{eq:avail} excludes non-executable candidates, and
\eqref{eq:closure} enforces predecessor closure.  A hard-budget variant
with $\sum_{i\in D^+(F)}c_i x_i\le\beta$ is a precedence-constrained
knapsack problem in general, so we use the fixed-$\lambda$ scalarized
selector as the default publication rule. We reserve \emph{budget} for this optional
hard cap $\beta$; all main experiments use the fixed-$\lambda$ selector and report the realized
normalized repair-operator \emph{Cost} of the executed plan.

\begin{theorem}[Scalarized repair via min-cut]
\label{thm:cost-aware-repair}
For fixed event $e$, candidate family
$Q_\rho=\{q_i:i\in D^+(F)\}$ with fixed $(P_i,v_i,w_i,c_i)$, and
$\lambda\ge0$, the scalarized repair-selection problem
\eqref{eq:scalarized-selection}--\eqref{eq:binary} reduces to
maximum-weight predecessor closure and can be solved exactly by a single
$s$--$t$ min-cut.
\end{theorem}

Briefly,
setting $p_i=w_i-\lambda c_i$ yields a maximum-weight closure problem:
positive nodes connect to the source, negative nodes connect to the sink,
and large-capacity edges enforce executability and predecessor closure.
Although $E^{\mathrm{req}}$ is oriented from prerequisite to dependent,
the min-cut network uses edges $i\to j$ for each $j\in\bar P_i$, so that
selecting $i$ forces selection of every required predecessor.



\begin{algorithm}[!ht]
\caption{\textsc{MemoRepair}}
\label{alg:memo-repair-main}
\small
\begin{algorithmic}[1]
\REQUIRE Repair event $e=(F,\tau,\Delta)$, graph $\mathcal{G}$, tradeoff $\lambda$
\ENSURE Republished validated successors $R_\rho$
\STATE Compute $\kappa_e$, $C(F)$, $D(F)$, and $\mathsf{Ret}(e)$
\STATE Set $B_\rho\leftarrow C(F)$ and withdraw $B_\rho$
\STATE Assign modes $\mu_i=\Phi_\mu^e(i;\mathcal G,\mathsf{Ret}(e),\kappa_e)$ for $i\in D(F)$
\STATE Construct $D^+(F)$, $Q_\rho$, $\{\bar P_i\}$, and $E^{\mathrm{req}}$
\STATE Mark unavailable, non-repairable, or cyclic candidates non-executable
\STATE Solve \eqref{eq:scalarized-selection}--\eqref{eq:binary} and set $S_\rho=\{i:x_i=1\}$
\STATE Execute $S_\rho$ in topological order over $E^{\mathrm{req}}[S_\rho]$
\STATE Republish validated successors; keep failures and their dependents withdrawn
\RETURN $R_\rho$
\end{algorithmic}
\end{algorithm}

\section{Experiments}
\label{sec:experiment}


\paragraph{Benchmarks.}
We evaluate \textsc{MemoRepair} on ToolBench and MemoryArena. \textbf{ToolBench}~\citep{qin2023toolllm}, executed via StableToolBench~\citep{guo2024stabletoolbench}, provides tool-use
trajectories with API calls, cached outputs, and executable procedures.
For parametric repair, we use the released ToolLLaMA-7B LoRA checkpoint
trained on ToolBench data as a ToolBench-derived neural artifact
\citep{toolllama_lora_hf}. \textbf{MemoryArena}~\citep{memoryarena2026} provides long-horizon
memory trajectories with preferences, episodic records, and retrieval
traces.

\textbf{Event and cascade statistics.}
Cascades range from single-tool deletions with a few dependents to
collection-level migrations affecting tens of artifacts, yielding
nondegenerate selection instances.

\paragraph{Metrics.}
We report exposure, successor publication, and downstream task effects.
\textbf{Leak\%} ($\downarrow$) is the fraction of affected artifacts in $C(F)$
that remain servable after repair. \textbf{Stale-use\%} ($\downarrow$) is the
fraction of post-event action traces that use invalidated information.
\textbf{Rep.\%} ($\uparrow$) is the validated republication rate over
non-remove repair contracts, micro-averaged over events as
$100\sum_e |R_{\rho,e}|/\sum_e |D^+_e(F)|$. \textbf{Cost} ($\downarrow$)
is normalized executed repair-operator cost, excluding the withdrawal barrier.
\textbf{$\Delta$Task} ($\uparrow$) is a diagnostic task score, reported in
percentage points. For task $t$, let $A_t$ be the affected artifacts used by
the task, $U_t\subseteq A_t$ those without a validated successor, and
$L_t\subseteq A_t$ those whose stale version remains visible. We set $s_t=1$
when $A_t=\emptyset$ and otherwise
$
s_t = 1-\frac{|U_t|+\frac12|L_t|}{|A_t|}.
$
We report $100(\mathbb{E}_t[s_t]-1)$. The sets $U_t$ and $L_t$ may overlap:
\emph{No action} can be penalized for both missing repair and stale serving,
whereas \emph{Remove all} is penalized only for missing repair. For neural
skills, \textbf{FSP\%}, \textbf{RU\%}, and \textbf{Val.\ pass\%} report
forget-set suppression, retained utility, and threshold-passing parametric
successors. 

\paragraph{Baselines.}
\emph{Internal repair policy baselines.}
(i)~\textsc{No action}: the store is left unchanged.
(ii)~\textsc{Remove all}: the affected cascade $C(F)$ is withdrawn,
no successors are republished, and no repair operators are executed.
(iii)~\textsc{Barrier+Greedy}: a selector-ablation control that
shares the withdrawal barrier, candidates, repair operators, and
validation oracle with \textsc{MemoRepair}, replacing only the exact
min-cut selector with a greedy value-to-cost heuristic under the same
$\lambda$. Differences against \textsc{Barrier+Greedy} are therefore
attributable to the selector, not to the contract.
(iv)~\textsc{Repair all}: executes every initially executable repair
candidate and republishes only successors that pass $\mathrm{Validate}_i$.
It therefore defines the publication ceiling rather than a $100\%$ publication
baseline; its Cost is $1.00$ whenever the initially executable set $E_0$ is
nonempty, and $0.00$ otherwise.

\emph{External memory systems baselines.}
We evaluate Mem0~\citep{mem0}, Zep/Graphiti~\citep{zep2025},
MemR$^{3}$~\citep{du2025memr}, TierMem~\citep{tiermem2025},
A-MEM~\citep{xu2025amem}, and O-Mem~\citep{omem2025} under their native
update and retrieval semantics. These systems do not expose cascade withdrawal, successor validation,
or predecessor-closed republication. We use them as stale-resistance reference points, measuring how much
invalidated content their native pipelines serve or retrieve after an
event on the same traces. 

\emph{Neural-skill operators.}
For the $\mathsf{param}$ repair mode, we compare adapted
SBU~\citep{wang2026agenticunlearning} and six parameter-level
operators (NPO~\citep{zhang2024npo}, SimNPO~\citep{simnpo2024},
BLUR~\citep{blur2026}, RMU~\citep{rmu2024}, LUNE~\citep{lune2025}, and
OOO~\citep{ooo2024}) inside the same \textsc{MemoRepair} pipeline,
using the same provenance-derived
$(\mathsf{For}_i(e),\mathsf{Ref}_i(e))$ partitions and validation oracle.

\subsection{Repair Policy Comparison}
\label{subsec:repair-policy}

Table~\ref{tab:internal_main} reports repair-policy comparisons on
ToolBench and MemoryArena. Under complete influence provenance, every
artifact in $C(F)$ becomes non-servable once the withdrawal barrier is
acknowledged, so zero Leak and Stale-use are contract guarantees for
all withdrawal-based methods. The empirical question reduces to
whether validated successor publication recovers task utility at lower
repair-operator cost than exhaustive repair. \emph{Remove all} isolates
the barrier by withdrawing the cascade without invoking any repair
operator, yielding Rep.\ $=$ Cost $=0$. \emph{Repair all} establishes
the publication ceiling: it attempts every initially executable
candidate and republishes only successors that pass $\mathrm{Validate}_i$.
\textsc{MemoRepair} and \textsc{Barrier+Greedy} share the barrier,
candidate construction, operators, and validation oracle, and differ
only in the publication selector. 


On ToolBench, \textsc{MemoRepair} republishes $22.4\%$, $45.1\%$, and
$36.5\%$ of $D^+(F)$ under deletion, correction, and migration
respectively, recovering $91.1$--$94.3\%$ of the \emph{Repair all}
Rep.\ ceiling at Cost between $0.57$ and $0.66$. Its $\Delta$Task lies
within $0.08$ task points of \emph{Repair all} on all three event
types. On MemoryArena, deletion leaves fewer repairable descendants.
Deterministic record and final-answer descendants usually require the
deleted parent as input, so they become structurally non-executable
once that parent is withdrawn. Skill descendants can still be repaired
by the parametric repair operator, which removes the deleted support
from the skill's retained reference set rather than recomputing from
the deleted parent.


The
resulting Rep.\ ceiling is $5.5\%$, of which \textsc{MemoRepair}
attains $5.1\%$ at Cost $0.62$, with $\Delta$Task within $0.05$ task
points of \emph{Repair all}. Correction and migration admit larger
repair surfaces: \textsc{MemoRepair} republishes $33.1\%$ and $17.0\%$
of $D^+(F)$ at Cost $0.76$ and $0.64$ respectively, again within
$0.07$ task points of the \emph{Repair all} ceiling.

\begin{table*}[!ht]
  \centering\small
  \setlength{\tabcolsep}{2pt}
  \begin{tabular}{l@{\hskip 2pt}rrrrr@{\hskip 4pt}rrrrr@{\hskip 4pt}rrrrr}
  \toprule
  \multirow{2}{*}{\textbf{Method}} &
  \multicolumn{5}{c}{\textbf{Deletion}} &
  \multicolumn{5}{c}{\textbf{Correction}} &
  \multicolumn{5}{c}{\textbf{Migration}} \\
  \cmidrule(lr){2-6}
  \cmidrule(lr){7-11}
  \cmidrule(lr){12-16}
  & Leak & Stale & Rep. & Cost & $\Delta$Task
  & Leak & Stale & Rep. & Cost & $\Delta$Task
  & Leak & Stale & Rep. & Cost & $\Delta$Task \\
  \midrule
  \multicolumn{16}{c}{\textbf{\textit{ToolBench}}} \\
  \midrule
  No action
  & 100 & 100 & -- & -- & $\text{-}2.78$
  & 100 & 100 & -- & -- & $\text{-}2.78$
  & 100 & 100 & -- & -- & $\text{-}5.61$ \\
  \rowcolor{gray!12} Remove all
  & 0 & 0 & 0.0 & 0.00 & $\text{-}1.96$
  & 0 & 0 & 0.0 & 0.00 & $\text{-}1.96$
  & 0 & 0 & 0.0 & 0.00 & $\text{-}4.47$ \\
  Repair all
  & 0 & 0 & \underline{\textbf{24.6}} & 1.00 & $\underline{\mathbf{\text{-}1.54}}$
  & 0 & 0 & \underline{\textbf{48.8}} & 1.00 & $\underline{\mathbf{\text{-}1.05}}$
  & 0 & 0 & \underline{\textbf{38.7}} & 1.00 & $\underline{\mathbf{\text{-}2.98}}$ \\
  \rowcolor{gray!12} Barrier+Greedy
  & 0 & 0 & 20.3 & 0.72 & $\text{-}1.68$
  & 0 & 0 & 41.2 & 0.62 & $\text{-}1.20$
  & 0 & 0 & 31.4 & 0.76 & $\text{-}3.28$ \\
  \rowcolor{blue!10} \textsc{MemoRepair}
  & 0 & 0 & 22.4 & \underline{\textbf{0.61}} & $\text{-}1.60$
  & 0 & 0 & 45.1 & \underline{\textbf{0.57}} & $\text{-}1.11$
  & 0 & 0 & 36.5 & \underline{\textbf{0.66}} & $\text{-}3.06$ \\
  \midrule
  \multicolumn{16}{c}{\textbf{\textit{MemoryArena}}} \\
  \midrule
  No action
  & 100 & 100 & -- & -- & $\text{-}4.44$
  & 100 & 100 & -- & -- & $\text{-}4.44$
  & 100 & 100 & -- & -- & $\text{-}3.86$ \\
  \rowcolor{gray!12} Remove all
  & 0 & 0 & 0.0 & 0.00 & $\text{-}4.00$
  & 0 & 0 & 0.0 & 0.00 & $\text{-}4.00$
  & 0 & 0 & 0.0 & 0.00 & $\text{-}3.58$ \\
  Repair all
  & 0 & 0 & \underline{\textbf{5.5}} & 1.00 & $\underline{\mathbf{\text{-}3.74}}$
  & 0 & 0 & \underline{\textbf{35.6}} & 1.00 & $\underline{\mathbf{\text{-}2.91}}$
  & 0 & 0 & \underline{\textbf{18.4}} & 1.00 & $\underline{\mathbf{\text{-}3.18}}$ \\
  \rowcolor{gray!12} Barrier+Greedy
  & 0 & 0 & 4.2 & 0.70 & $\text{-}3.90$
  & 0 & 0 & 28.6 & 0.83 & $\text{-}3.17$
  & 0 & 0 & 14.2 & 0.71 & $\text{-}3.34$ \\
  \rowcolor{blue!10} \textsc{MemoRepair}
  & 0 & 0 & 5.1 & \underline{\textbf{0.62}} & $\text{-}3.79$
  & 0 & 0 & 33.1 & \underline{\textbf{0.76}} & $\text{-}2.98$
  & 0 & 0 & 17.0 & \underline{\textbf{0.64}} & $\text{-}3.23$ \\
  \bottomrule
  \end{tabular}
    \caption{Internal repair policy comparison on ToolBench and
MemoryArena. Underlines mark the column best among methods that
produce validated repairs.}
  \label{tab:internal_main}
  \end{table*}

\textbf{Selector frontier.}
Figure~\ref{fig:mincut_pareto} traces the repair--cost frontier obtained
by sweeping $\lambda$ in the scalarized min-cut selector on ToolBench
across deletion, correction, and migration. Each curve reports the
mean over three seeds, with shaded bands and horizontal caps showing
$\pm 1$ std on Rep and Cost respectively. The endpoints recover
\emph{Remove all} at $\lambda \to \infty$ and \emph{Repair all} at
$\lambda = 0$, and increasing $\lambda$ shifts the operating point
toward lower cost and lower validated publication. The configuration
used in Table~\ref{tab:internal_main} fixes $\lambda = 0.3$, which
sits at the knee of all three curves: the marginal slope
$\Delta\mathrm{Rep}/\Delta\mathrm{Cost}$ falls by a factor of four to
seven past this point while still recovering $91.1$--$94.3\%$ of the
\emph{Repair all} ceiling at Cost between $0.57$ and $0.66$.




\subsection{Cascade-Unaware Memory System Comparison}
\label{subsec:native-memory}

Table~\ref{tab:external_main} evaluates six external memory systems
under their own update and retrieval semantics. These systems are
cascade-unaware: none of them implements withdrawal barriers,
validated successor publication, or predecessor-closed repair
selection. They therefore serve as stale-resistance reference points
rather than repair-policy baselines. A value of $0$ in Rep.\ indicates
that the system performs no validated successor-publication operation
rather than a failed repair attempt; Cost is undefined because these
systems do not execute the repair operators associated with the $c_i$
terms in our selector.

Across ToolBench and MemoryArena the six systems reduce exposure
relative to \emph{No action}, yet a large fraction of stale
descendants remains servable. On ToolBench they leave $69.8\%$ to
$93.1\%$ of the affected cascade leakable and $92.4\%$ to $99.7\%$ of
post-event actions stale; on MemoryArena the corresponding ranges
are $73.9\%$ to $94.3\%$ and $93.2\%$ to $99.6\%$. The associated
$\Delta$Task remains within $0.24$ points of \emph{No action} on
ToolBench and within $0.37$ points on MemoryArena, indicating that
retrieval freshness alone does not provide withdrawal or validated
successor semantics. The \textsc{MemoRepair} row is included as an explicit cascade-repair
reference under complete influence provenance. 


\begin{table*}[htbp]
  \centering\small
  \setlength{\tabcolsep}{2pt}
  \begin{tabular}{l@{\hskip 2pt}rrrrr@{\hskip 4pt}rrrrr@{\hskip 4pt}rrrrr}
  \toprule
  \multirow{2}{*}{\textbf{System}} &
  \multicolumn{5}{c}{\textbf{Deletion}} &
  \multicolumn{5}{c}{\textbf{Correction}} &
  \multicolumn{5}{c}{\textbf{Migration}} \\
  \cmidrule(lr){2-6}
  \cmidrule(lr){7-11}
  \cmidrule(lr){12-16}
  & Leak & Stale & Rep. & Cost & $\Delta$Task
  & Leak & Stale & Rep. & Cost & $\Delta$Task
  & Leak & Stale & Rep. & Cost & $\Delta$Task \\
  \midrule
  \multicolumn{16}{c}{\textbf{\textit{ToolBench}}} \\
  \midrule
  No action
  & 100 & 100 & -- & -- & $\text{-}2.78$
  & 100 & 100 & -- & -- & $\text{-}2.78$
  & 100 & 100 & -- & -- & $\text{-}5.61$ \\
  \rowcolor{gray!12} Mem0
  & 74.6 & 93.8 & 0 & -- & $\text{-}2.61$
  & 80.2 & 95.6 & 0 & -- & $\text{-}2.66$
  & 91.3 & 98.9 & 0 & -- & $\text{-}5.46$ \\
  Zep/Graphiti
  & 69.8 & 92.4 & 0 & -- & $\text{-}2.54$
  & 76.5 & 94.1 & 0 & -- & $\text{-}2.59$
  & 89.1 & 98.4 & 0 & -- & $\text{-}5.39$ \\
  \rowcolor{gray!12} MemR$^{3}$
  & 85.2 & 98.6 & 0 & -- & $\text{-}2.72$
  & 87.4 & 98.9 & 0 & -- & $\text{-}2.74$
  & 92.6 & 99.6 & 0 & -- & $\text{-}5.48$ \\
  A-MEM
  & 72.1 & 92.7 & 0 & -- & $\text{-}2.56$
  & 79.4 & 95.0 & 0 & -- & $\text{-}2.63$
  & 90.6 & 98.7 & 0 & -- & $\text{-}5.42$ \\
  \rowcolor{gray!12} O-Mem
  & 88.7 & 99.4 & 0 & -- & $\text{-}2.73$
  & 88.9 & 99.6 & 0 & -- & $\text{-}2.74$
  & 93.1 & 99.7 & 0 & -- & $\text{-}5.51$ \\
  TierMem
  & 88.4 & 99.2 & 0 & -- & $\text{-}2.72$
  & 88.6 & 99.5 & 0 & -- & $\text{-}2.74$
  & 93.0 & 99.7 & 0 & -- & $\text{-}5.50$ \\
  \rowcolor{blue!10} \textsc{MemoRepair}
  & 0 & 0 & 22.4 & 0.61 & $\underline{\mathbf{\text{-}1.60}}$
  & 0 & 0 & 45.1 & 0.57 & $\underline{\mathbf{\text{-}1.11}}$
  & 0 & 0 & 36.5 & 0.66 & $\underline{\mathbf{\text{-}3.06}}$ \\
  \midrule
  \multicolumn{16}{c}{\textbf{\textit{MemoryArena}}} \\
  \midrule
  No action
  & 100 & 100 & -- & -- & $\text{-}4.44$
  & 100 & 100 & -- & -- & $\text{-}4.44$
  & 100 & 100 & -- & -- & $\text{-}3.86$ \\
  \rowcolor{gray!12} Mem0
  & 78.6 & 94.8 & 0 & -- & $\text{-}4.31$
  & 82.4 & 96.1 & 0 & -- & $\text{-}4.18$
  & 91.7 & 98.0 & 0 & -- & $\text{-}3.72$ \\
  Zep/Graphiti
  & 73.9 & 93.2 & 0 & -- & $\text{-}4.25$
  & 78.1 & 94.6 & 0 & -- & $\text{-}4.07$
  & 88.4 & 97.3 & 0 & -- & $\text{-}3.65$ \\
  \rowcolor{gray!12} MemR$^{3}$
  & 86.7 & 98.3 & 0 & -- & $\text{-}4.39$
  & 88.2 & 98.7 & 0 & -- & $\text{-}4.31$
  & 93.5 & 99.1 & 0 & -- & $\text{-}3.79$ \\
  A-MEM
  & 75.4 & 93.6 & 0 & -- & $\text{-}4.27$
  & 80.3 & 95.2 & 0 & -- & $\text{-}4.12$
  & 90.5 & 97.6 & 0 & -- & $\text{-}3.68$ \\
  \rowcolor{gray!12} O-Mem
  & 89.6 & 99.5 & 0 & -- & $\text{-}4.42$
  & 90.1 & 99.6 & 0 & -- & $\text{-}4.36$
  & 94.3 & 99.5 & 0 & -- & $\text{-}3.83$ \\
  TierMem
  & 89.2 & 99.3 & 0 & -- & $\text{-}4.41$
  & 89.8 & 99.5 & 0 & -- & $\text{-}4.35$
  & 93.9 & 99.4 & 0 & -- & $\text{-}3.82$ \\
  \rowcolor{blue!10} \textsc{MemoRepair}
  & 0 & 0 & 5.1 & 0.62 & $\underline{\mathbf{\text{-}3.79}}$
  & 0 & 0 & 33.1 & 0.76 & $\underline{\mathbf{\text{-}2.98}}$
  & 0 & 0 & 17.0 & 0.64 & $\underline{\mathbf{\text{-}3.23}}$ \\
  \bottomrule
  \end{tabular}
\caption{Stale-resistance comparison with memory systems without cascade
repair. For external systems, Rep.\ $=0$ means no validated successor
publication and Cost is undefined; \textsc{MemoRepair} is included only as a
cascade repair reference.}
  \label{tab:external_main}
\end{table*}

\subsection{Parametric Repair}
\label{subsec:neural}

\begin{figure*}[htbp]
    \centering
    \includegraphics[width=\linewidth]{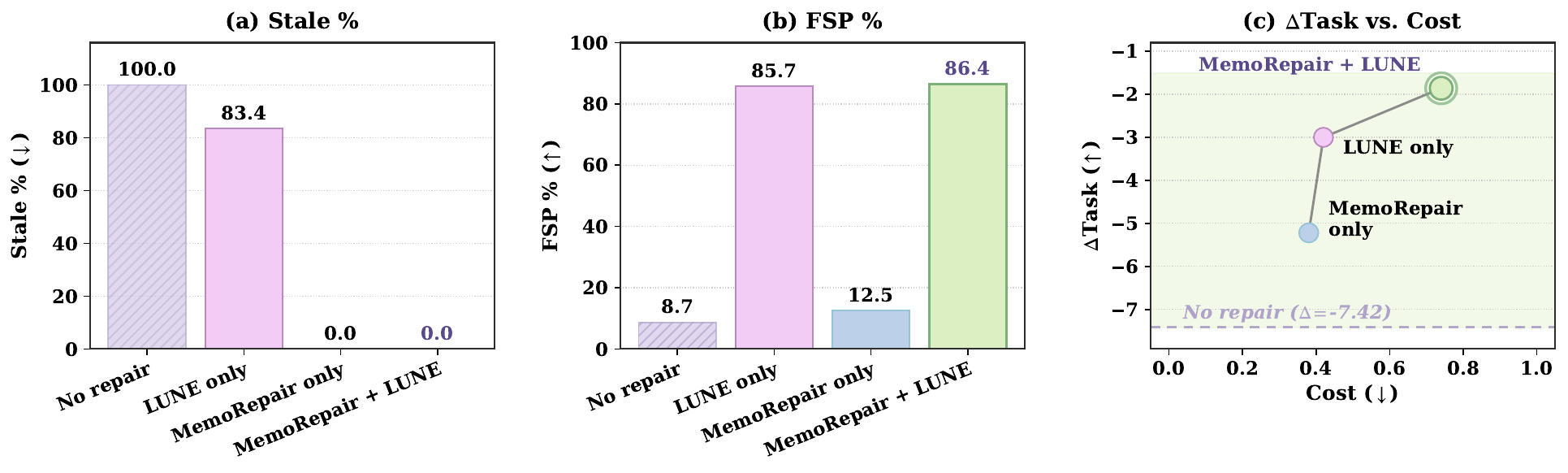}
    \caption{Parametric-repair pipeline ablation on the ToolBench neural-skill subset.}
    \label{fig:parametric_pipeline}
\end{figure*}

Figure~\ref{fig:parametric_pipeline} ablates the two repair levels.
Without repair, the LoRA checkpoint has $\text{Stale}=100$,
$\text{FSP}=8.7$, and $\Delta\text{Task}=-7.42$, indicating both
visible stale serving and parametric persistence of invalidated
support. Param-only LUNE raises FSP to $85.7$ and recovers
$\Delta\text{Task}$ to $-3.00$, but leaves Stale at $83.4$ because
materialized descendants remain visible. Cascade-only
\textsc{MemoRepair} drives Stale to $0$ at $\Delta\text{Task}=-5.22$,
yet leaves FSP near baseline at $12.5$ because model weights are not
updated. The two repair levels fix different failure modes: parameter-side
repair reduces stale influence inside the neural skill, while
cascade-side repair prevents stale materialized artifacts from
remaining visible. Composing them reaches $\text{Stale}=0$,
$\text{FSP}=86.4$, and $\Delta\text{Task}=-1.86$, improving
$\Delta\text{Task}$ by $5.56$ points over no repair and by $1.14$
and $3.36$ points over LUNE only and cascade only respectively. The
composed run is costed as a single pipeline, so shared replay or
sandbox work is charged once; its Cost is $0.74$, below the
independent-run sum $0.42+0.38=0.80$.

\begin{figure*}[htbp]
\centering
\begin{minipage}[l]{0.5\linewidth}
\centering\small
\setlength{\tabcolsep}{3.2pt}
\begin{tabular}{lcccc}
\toprule
\textbf{Operator} &
\textbf{FSP\%}$\uparrow$ &
\textbf{RU\%}$\uparrow$ &
\textbf{Val.\,pass\%}$\uparrow$ &
\textbf{Rel.\,comp.}$\downarrow$ \\
\midrule
Full retrain
  & 91.8 & 96.2 & 92.5 & $8.4\times$ \\
SBU    & 86.9 & 92.8 & 83.6 & $1.8\times$ \\
NPO    & 81.7 & 88.9 & 72.4 & $1.3\times$ \\
SimNPO & 80.4 & 90.7 & 75.8 & $1.2\times$ \\
RMU    & 78.9 & 91.1 & 71.6 & $1.5\times$ \\
BLUR   & \underline{\textbf{88.2}} & 91.9 & 82.1 & $3.7\times$ \\
LUNE   & 86.4 & \underline{\textbf{93.4}} & \underline{\textbf{84.9}} & $\underline{\mathbf{1.0\times}}$ \\
OOO    & 87.4 & 92.5 & 83.2 & $1.4\times$ \\
\bottomrule
\end{tabular}
\captionof{table}{Parameter-level repair operators inside
\textsc{MemoRepair}. Full retrain uses retained support
$\mathsf{Ref}_i(e)$. Rel.\ comp.\ is wall-clock compute normalized to LUNE,
not the Cost metric.}
\label{tab:parametric_operators}
\end{minipage}\hfill
\begin{minipage}[c]{0.45\linewidth}
\centering
\includegraphics[width=\linewidth]{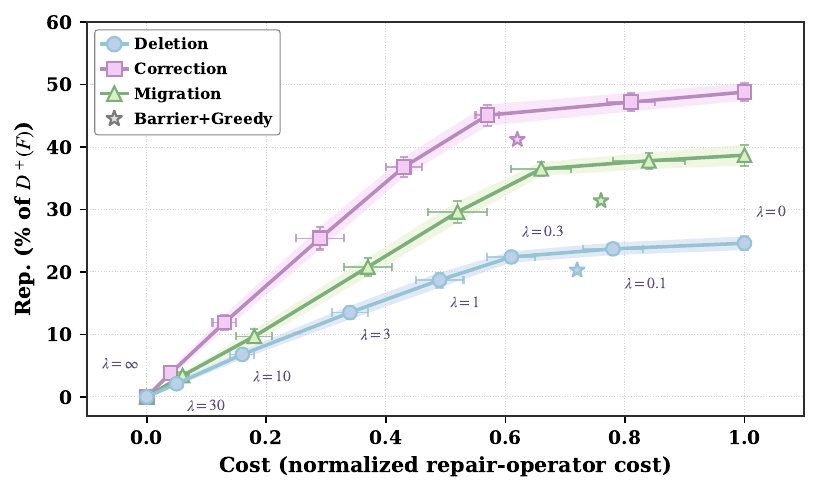}
\captionof{figure}{Min-cut repair--cost frontier on ToolBench, traced by sweeping $\lambda$.}
\label{fig:mincut_pareto}
\end{minipage}
\end{figure*}

Table~\ref{tab:parametric_operators} compares parameter-level operators
inside the same \textsc{MemoRepair} pipeline. All operators use the same
$(\mathsf{For}_i(e),\mathsf{Ref}_i(e))$ partitions and validation oracle.
Full retraining reaches $\text{FSP}=91.8$ and
$\text{Val.\,pass}=92.5$ at $8.4\times$ relative compute, serving as a
high-compute quality ceiling. Among practical operators, LUNE gives
the best observed compute--quality tradeoff: it has the highest RU
($93.4$) and Val.\ pass ($84.9$), while BLUR attains the highest FSP
($88.2$) at $3.7\times$ relative compute. NPO, SimNPO, and RMU trail
in this setting ($\text{FSP}<82$, $\text{Val.\,pass}<76$). We use
LUNE as the default $\mathsf{param}$ operator in \textsc{MemoRepair};
BLUR is a higher-compute alternative when maximizing FSP is the
priority.

\subsection{Ablations and Robustness}
\label{subsec:ablations_robustness}


\begin{table}[!ht]
\centering\small
\setlength{\tabcolsep}{3.4pt}
\begin{tabular}{llccc}
\toprule
\textbf{Type} & \textbf{Setting} &
\textbf{Leak\%}$\downarrow$ &
\textbf{Stale-use\%}$\downarrow$ &
\textbf{Repair time rel.} \\
\midrule
\multirow{2}{*}{Withdrawal}
& No barrier
  & 100.0 & 100.0 & $0.52\times$ \\
& \textbf{\textsc{MemoRepair}}
  & \underline{\textbf{0.0}} & \underline{\textbf{0.0}} & $1.00\times$ \\
\midrule
\multirow{3}{*}{Validation}
& Schema-only
  & 0.0 & 88.6 & -- \\
& Task-regression-only
  & 0.0 & 29.8 & -- \\
& \textbf{Full $\mathrm{Validate}_i$}
  & \underline{\textbf{0.0}} & \underline{\textbf{0.0}} & -- \\
\midrule
\multirow{4}{*}{Provenance}
& $p_{\mathrm{drop}}=0.005$
  &  8.6 &  9.4 & -- \\
& $p_{\mathrm{drop}}=0.010$
  & 17.7 & 19.7 & -- \\
& $p_{\mathrm{drop}}=0.020$
  & 34.2 & 38.0 & -- \\
& $p_{\mathrm{drop}}=0.050$
  & 61.8 & 68.5 & -- \\
\midrule
Concurrency
& Overlapping events
  & 0.0 & 0.0 & $1.86\times$ \\
Replica
& Partition + rejoin
  & 0.0 & 0.0 & $1.12\times$ \\
\bottomrule
\end{tabular}
\caption{Component ablations and robustness checks. Withdrawal rows
report repair-interval exposure (not directly comparable to the
post-repair Leak/Stale in
Tables~\ref{tab:internal_main}--\ref{tab:external_main}); Validation
rows report the oracle's false-pass rate in the Stale-use column.
Provenance, Concurrency, and Replica share the \textsc{MemoRepair}
baseline.}
\label{tab:ablations}
\end{table}

Table~\ref{tab:ablations} reports component ablations and robustness
checks. The withdrawal barrier trades a $0.52\times$ time saving for
zero stale exposure during repair: removing it leaves the cascade
servable throughout ($\text{Leak}=\text{Stale}=100$), an
availability/correctness choice that no barrier-free design can
avoid. The validation ablation isolates two failure modes:
schema-only and task-regression-only validation each miss what the
other catches ($88.6\%$ and $29.8\%$ stale-use), and only the
composed oracle reaches zero. Provenance fidelity is leveraged:
dropping $1\%$ of influence edges yields $17.7\%$ Leak ($\approx
18\times$ amplification), a graceful linear degradation that
identifies complete influence provenance as the main system-level
invariant behind the contract guarantee. Overlapping events and
partition+rejoin add $1.86\times$ and $1.12\times$ repair time while
preserving $\text{Leak}=\text{Stale}=0$.

\section{Related Work}
\label{sec:related}

\paragraph{Agent memory and tool-derived state.}
LLM agents increasingly maintain state beyond the immediate context window. 
Generative Agents and Reflexion store memories or reflections for later decisions, 
while MemGPT and A-MEM study mechanisms for managing long-term agent memory 
across interactions~\citep{park2023generative,shinn2023reflexion,
packer2023memgpt,xu2025amem}. Skill-oriented agents further show that past
experience can be distilled into reusable procedures or skills~\citep{mi2026procmem}.
A related source of persistent state appears in tool-use systems, where API calls,
arguments, outputs, cached responses, and traces become reusable workflow artifacts
~\citep{qin2023toolllm,guo2024stabletoolbench}. These works motivate the memory
substrate considered here. \textsc{MemoRepair} studies a later stage in the lifecycle:
after an artifact has already been materialized, a deletion, correction, or migration
can make its descendants stale, requiring explicit withdrawal and repair.

\paragraph{Unlearning and memory privacy.}
Machine unlearning removes or suppresses the influence of deleted data in learned
models~\citep{cao2015towards,guo2020certified,xu2024unlearning}. Recent work
extends forgetting to agent settings, where information may persist in behavior,
retrieval, or memory access~\citep{wang2026agenticunlearning}. Privacy studies of
LLM-agent memory similarly show that stored interactions can remain exposed through
memory retrieval~\citep{wang2025mextra}. These lines address model-side, retriever-side,
or access-side influence. \textsc{MemoRepair} addresses a separate store-level failure
mode: summaries, cached outputs, prompt skills, chain procedures, and neural-skill
artifacts may remain visible even after the source artifact has been updated or removed.

\paragraph{Provenance, view maintenance, and selection.}
Database provenance and view maintenance provide tools for relating source data to
derived state and propagating updates through materialized views
~\citep{buneman2001prov,cheney2009provenance,ceri1991deriving}. Agent memory
requires additional machinery because descendants are heterogeneous artifacts, repair
operators may fail validation, and publishing one successor may require other repaired
successors to be available first. Our selector uses the classical maximum-closure/min-cut
connection~\citep{picard1976closure}. The contribution is not a new graph-cut algorithm,
but the cascade-repair contract that turns successor publication in agent memory into a
predecessor-closed selection problem.

\section{Limitations and Conclusion}
\label{sec:conclusion}

\paragraph{Limitations.}
\textsc{MemoRepair} relies on complete influence provenance: each durable
artifact must record the artifacts used to derive it. Missing edges can leave
affected descendants outside the withdrawn cascade. This assumption is visible
in Table~\ref{tab:ablations}: dropping $1\%$ of influence edges yields
$17.7\%$ Leak, with similar scaling for small $p_{\mathrm{drop}}$. The method
also depends on the coverage of the validation suite. A successor that passes
the implemented checks is not guaranteed to be semantically correct in all
future contexts. For neural skills, \textsc{MemoRepair} inherits the guarantees
and failure modes of the chosen parameter-level repair operator; the store-level
contract should therefore not be read as a claim of exact parameter erasure.

\paragraph{Conclusion.}
We studied cascade repair for the visible derived state of agentic memory.
When a source artifact is deleted, corrected, or migrated, downstream summaries,
cached outputs, procedures, and skills may remain servable with stale support.
\textsc{MemoRepair} addresses this post-update state by withdrawing the affected
cascade, constructing successors from retained support and staged repaired
predecessors, and republishing only successors that pass validation and satisfy
predecessor closure. For a fixed repair--cost tradeoff, the publication problem
reduces to maximum-weight predecessor closure and is solved by a single min-cut.
Under complete influence provenance, the withdrawal barrier gives store-level
stale-serving isolation, while the selector recovers most of the repair-all
publication ceiling at lower repair-operator cost.

\bibliographystyle{plainnat}
\bibliography{related}



\end{document}